\documentclass{article}

\usepackage[preprint,nonatbib]{neurips_2024}

\usepackage[pagebackref,breaklinks,colorlinks]{hyperref}
\usepackage[utf8]{inputenc} 
\usepackage[T1]{fontenc}    
\usepackage{hyperref}       
\usepackage{url}            
\usepackage{booktabs}       
\usepackage{amsfonts}       
\usepackage{nicefrac}       
\usepackage{microtype}      
\usepackage{xcolor}         
\usepackage{multirow}
\usepackage[pdftex]{graphicx}
\usepackage{amsmath}
\usepackage{algorithm}
\usepackage{algorithmic}
\usepackage{bbding}
\usepackage{pifont}
\usepackage{makecell}
\usepackage{subfig}

\title{Character-Adapter: Prompt-Guided Region Control for High-Fidelity Character Customization}

\author{
Yuhang Ma\textsuperscript{1}$^\ast$\thanks{Equal Contribution}\thanks{Project Lead},
~~~ Wenting Xu\textsuperscript{1}$^\ast$,
~~~ Jiji Tang\textsuperscript{1}$^\ast$,
~~~ Qinfeng Jin\textsuperscript{1},\\
~~~ \bf{Rongsheng Zhang}\textsuperscript{1},
~~~ \bf{Zeng Zhao}\textsuperscript{1}\thanks{Corresponding Authors},
~~~ \bf{Changjie Fan}\textsuperscript{1},
~~~ \bf{Zhipeng Hu}\textsuperscript{1}
\\
\textsuperscript{1}Fuxi AI Lab, NetEase Inc.\\
}

\begin{document}

\maketitle
\begin{abstract}
   
   Customized image generation, which seeks to synthesize images with consistent characters, holds significant relevance for applications such as storytelling, portrait generation, and character design. However, previous approaches have encountered challenges in preserving characters with high-fidelity consistency due to inadequate feature extraction and concept confusion of reference characters. Therefore, we propose \textbf{\emph{Character-Adapter}}, a plug-and-play framework designed to generate images that preserve the details of reference characters, ensuring high-fidelity consistency. Character-Adapter employs \textit{prompt-guided segmentation} to ensure fine-grained regional features of reference characters and \textit{dynamic region-level adapters} to mitigate concept confusion. Extensive experiments are conducted to validate the effectiveness of Character-Adapter. Both quantitative and qualitative results demonstrate that Character-Adapter achieves the state-of-the-art performance of consistent character generation, with an improvement of 24.8\% compared with other methods. Our code will be released at \url{https://github.com/Character-Adapter/Character-Adapter}.

\end{abstract}    
\section{Introduction}
\begin{figure*} [htbp]
\centering
\includegraphics[width=1\linewidth]{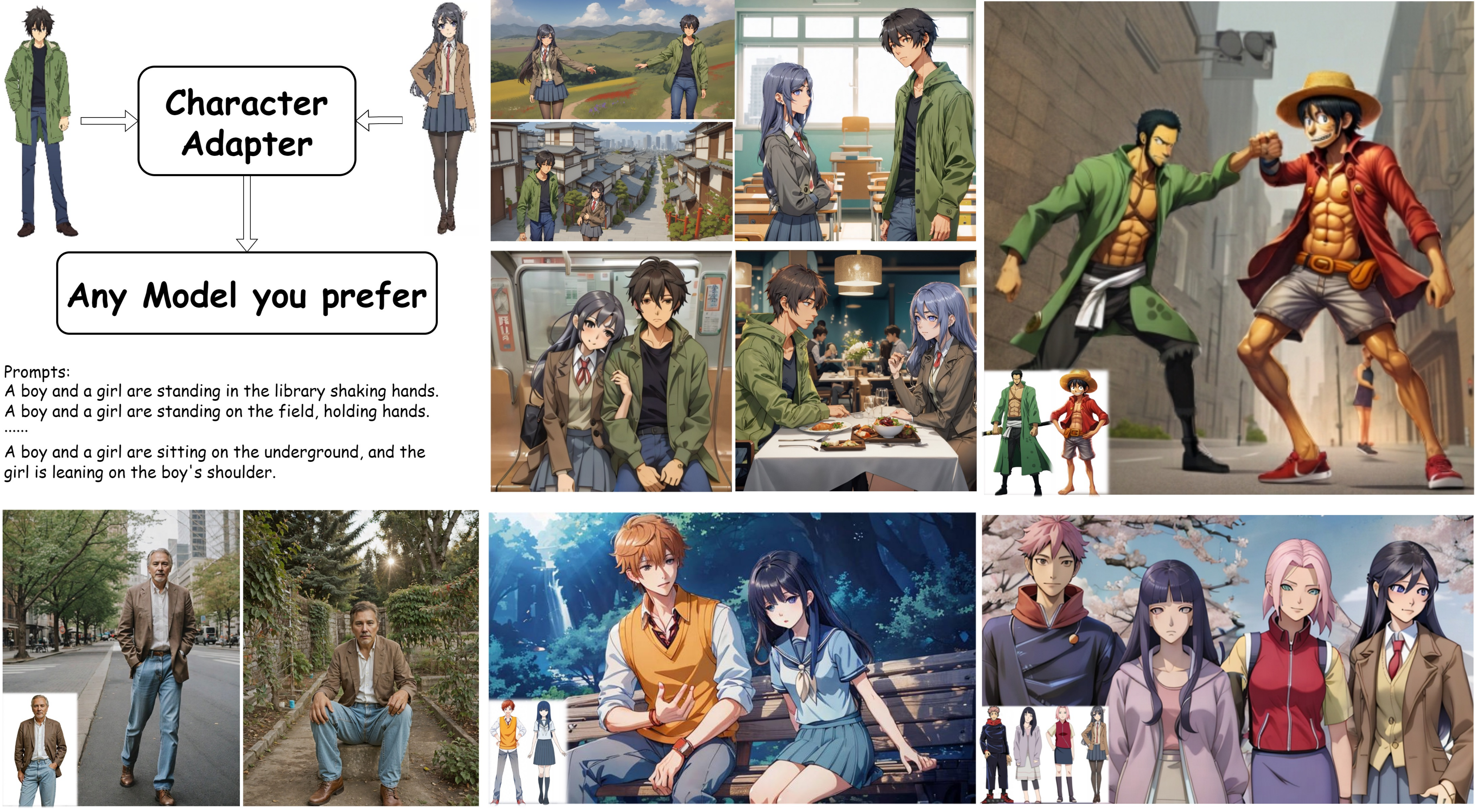}

\caption{\label{fig:intro} Images generated by Character-Adapter. Character-Adapter can be seamlessly integrated with any preferred model, without extra training. This approach empowers the customization of concepts while preserving the high-fidelity appearance of given characters (without any quantitative limitations), encompassing attributes such as hairstyle, identity, attire, and others.} 
\end{figure*}

Under the nurturing of text-to-image diffusion models, customized image generation aiming to synthesize images with consistent characters, holds significant relevance for applications such as storytelling, portrait generation, and character design.  Particularly, training-based methods \cite{ruiz2023dreambooth,hua2023dreamtuner,hu2021lora,wei2023elite,kumari2023multi,kong2024omg,gu2024mix} are the most prevalent approach for generating high-fidelity characters (e.g., Dreambooth \cite{ruiz2023dreambooth}, Custom Diffusion \cite{kumari2023multi}). 
However, these training-based methods exhibit several notable limitations. Firstly, they necessitate the acquisition of customized data to fine-tune the models, resulting in a significant demand for computational resources and prolonged training durations. Secondly, they may encounter trade-offs between character consistency and text-image alignment \cite{hu2021lora,gu2024mix,kong2024omg}. Furthermore, fine-tuning for specific characters can potentially compromise the robustness, leading to a sacrifice in their ability to generalize across a wide range of characters \cite{tewel2024training}. 

To inspire generalizability, training-free methods \cite{ye2023ip,wang2024instantid,li2023photomaker,xiao2023fastcomposer,li2024blip} prioritize consistent character generation by utilizing adapter modules or optimizing the parameters of diffusion models to incorporate reference images. They suffer from several drawbacks: 1) they primarily focus on identity preservation (e.g., IP-Adapter \cite{ye2023ip}, InstantID \cite{wang2024instantid}), neglecting other aspects of the reference character, such as attire and decorations \cite{xiao2023fastcomposer,wu2023easyphoto, wang2024instantid}; 2) they struggle in capturing precise semantic representations, as they primarily focus on the reference image \cite{ye2023ip,wang2024instantid}. 

We consider that the inconsistency generation exhibited by these methods is attributed to inadequate image feature extraction and concept confusion of reference characters \cite{gu2024mix}. Recently, IP-Adapter \cite{ye2023ip} is introduced to embed reference image features. However, integrating the entire reference image into one adapter can lead to inadequate feature preservation. Furthermore, given that the text encoder struggles to encapsulate compositional concepts, employing full tokens for multi-concept generation may lead to concept confusion \cite{dahary2024yourself}. To address these problems, We present Character-Adapter, a novel framework designed to facilitate consistent character generation. Specifically, it incorporates a \emph{\textbf{prompt-guided segmentation}} module that localizes image regions based on text prompts, thereby facilitating adequate image feature extraction. Subsequently, we introduce a \emph{\textbf{dynamic region-level adapters}} module, comprising region-level adapters and attention dynamic fusion. 
The region-level adapters allow each adapter to concentrate on the corresponding region (e.g., attire, decorations) of the generated image, thereby mitigating concept fusion and promoting disentangled representations for each component of the generated image. Additionally, the attention dynamic fusion is introduced to enable more accurate conditional image feature preservation while facilitating coherent generation between the character and the background regions. 

Overall, Character-Adapter is a plug-and-play framework designed to generate highly consistent characters with intricate details. Its advantage of not necessitating further training allows it to be more versatile and practical in its applications. Several cases 
involving both single and multiple character generation are illustrated in Fig. \ref{fig:intro}. Our contributions can be summarized as follows:

\begin{itemize}
    \item We introduce Character-Adapter, a framework designed to ensure high-fidelity character generation. Our method achieves the state-of-the-art performance of consistent character generation, with an improvement of 24.8\% compared with other methods.
    \item We propose prompt-guided segmentation for regional localization of reference characters to facilitate comprehensive feature extraction, and employ dynamic region-level adapters to mitigate concept fusion, thereby preserving high-fidelity consistency with reference characters.
    \item Character-Adapter is a versatile plug-and-play model that can be easily integrated into any backbone model or compatible with other editing tools (such as ControlNet \cite{zhang2023adding}) for both single and multiple character generation.
\end{itemize}
\section{Related work}

\textbf{Text-to-image generative models.} Diffusion models have achieved remarkable results in text-to-image generation in recent years \cite{nichol2021glide,saharia2022photorealistic,ramesh2022hierarchical,ho2020denoising,song2020denoising,balaji2022ediff,ramesh2021zero}. Early works such as DALL-E2 \cite{ramesh2022hierarchical} and Imagen \cite{saharia2022photorealistic} utilize original images as the diffusion input, resulting in enormous computational resources and training time. Latent diffusion models (\textbf{LDMs}) ~\cite{Rombach_2022_CVPR} have been introduced to compress images into a latent space through a pre-trained auto-encoder \cite{van2017neural}, instead of operating directly in the pixel space \cite{saharia2022photorealistic,nichol2021glide}. However, general diffusion models rely solely on text prompts, lacking the capability to generate consistent characters with image conditions.  

\textbf{Consistent character generation.} Subject-driven image generation aims to generate customized images of a particular subject based on different text prompts. Most existing works adopt extensive fine-tuning for each subject \cite{ruiz2023dreambooth,hua2023dreamtuner,hu2021lora,wei2023elite}. Dreambooth \cite{ruiz2023dreambooth} maps the subject to a unique identifier while Textual-Inversion \cite{gal2022textual} is proposed to optimize a word vector for a custom concept. 
Moreover, some works \cite{kumari2023multi,kong2024omg,gu2024mix} put their effort in multi-subject image generation. 
Custom Diffusion \cite{kumari2023multi} propose to combine multiple concepts via closed-form constrained optimization. OMG \cite{kong2024omg} and Mix-of-Show \cite{gu2024mix} propose to optimize the fusion mode during training in circumstance of multi-concept generation. However, these methods necessitate additional training for all subjects, which can be time-consuming in multi-subject generation scenarios. Recently, some methods strive to enable subject-driven image generation without additional training \cite{xiao2023fastcomposer,zhang2023ssr,ye2023ip,li2023photomaker,wang2024instantid,tewel2024training}. Most of them explore extended-attention mechanisms for maintaining identity consistency. IP-Adapter \cite{ye2023ip} and InstantID \cite{wang2024instantid} introduce visual control by separating cross-attention layers for text features and image features. ConsiStory \cite{tewel2024training} enables training-free subject-level consistency across novel images via cross-frame attention. However, they fail to preserve detailed information according to the inadequate image feature extraction.
\section{Methods}

\subsection{Preliminaries}

\textbf{Cross-attention in LDMs.} Cross-attention mechanisms are used to augment the underlying U-Net backbone, thereby transforming DMs into more flexible conditional image generators, which are effective for learning attention-based models for a variety of input modalities \cite{vaswani2017attention}. Specifically, the output after the cross-attention mechanism in the Stable Diffusion are noted as:
\begin{equation}
\label{eq:crossattention}
\begin{split}
{z}'=\text{Attention}({Q},{K},{V}) = \text{Softmax}(\frac{{Q}{K}^{\top}}{\sqrt{d}}){V},
\\
\end{split}
\end{equation}
where $Q = zW_{q}$,$K = c_{t}W_{k}$,$V = c_{t}W_{v}$, $z$ denotes latent vectors obtained from the encoder $\mathcal E$, $c_{t}$ denotes text features obtained from text encoder.
\begin{figure*} [t]
\centering
\includegraphics[width=1\linewidth]{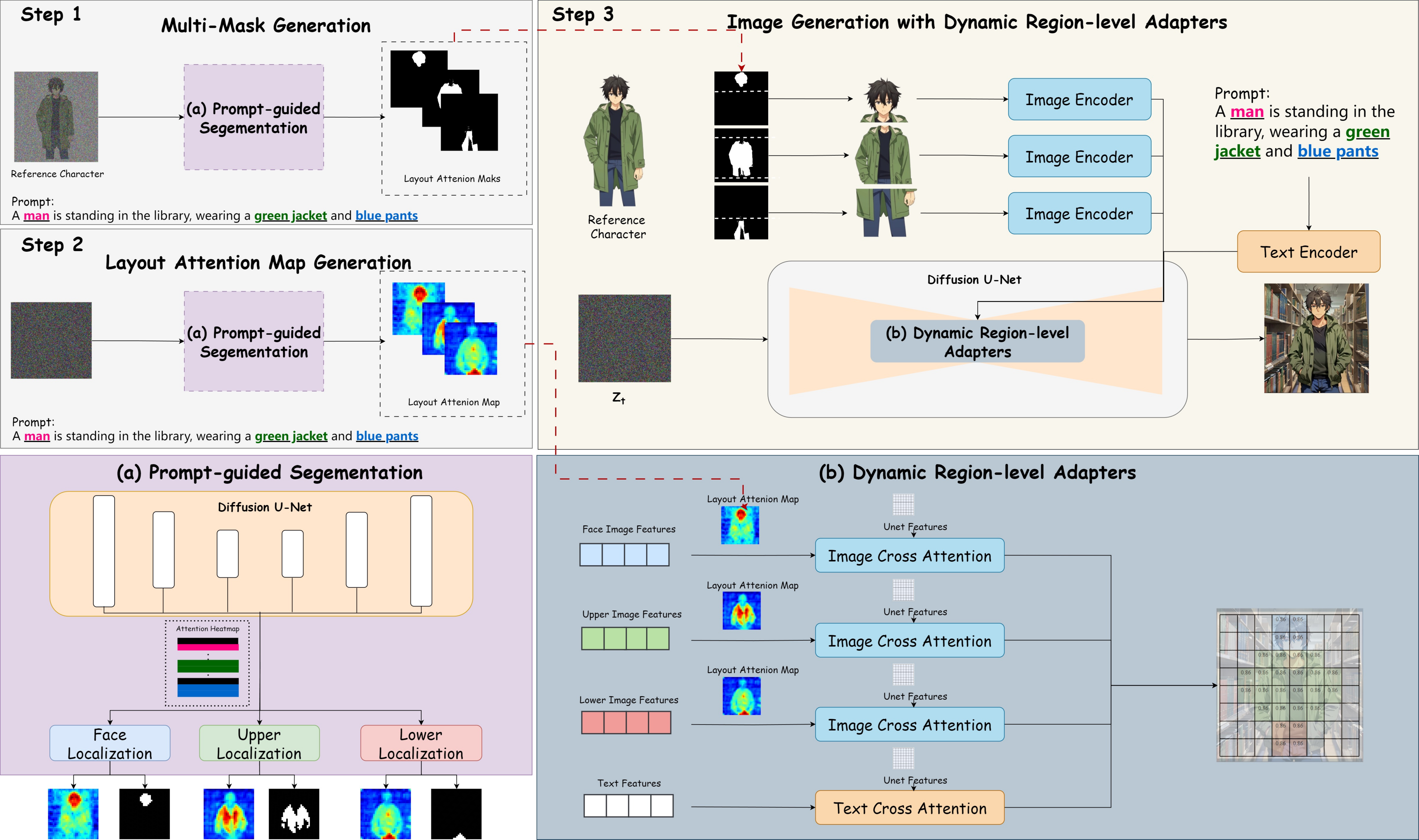} 
\vspace{-5pt}
\caption{\label{fig:overview1} Framework of Character-Adapter. Step 1 involves obtaining the segmentation of the reference characters with given images and prompts through the \textbf{prompt-guided segmentation} module (Module a). Step 2 acquires attention maps of layout images generated solely from the given prompts via the same module. Step 3 illustrates the process of generating images with the given prompt and semantic regions through the \textbf{dynamic region-level adapters} module (Module b). }
\vspace{-5pt}
\end{figure*} 
\textbf{Image Prompt Adapter.} IP-Adapter proposes a decoupled cross-attention strategy to support conditional image generation by introducing an image cross-attention mechanism \cite{ye2023ip} analogous to the original cross-attention module in Stable Diffusion. This mechanism seamlessly integrates image prompts with text prompts to guide the text-to-image generation process. The decoupled cross-attention can be formulated as:

\begin{equation}
\label{eq:ipadapter}
{Z}^{new}=\text{Attention}({Q},{K},{V}) + \lambda\cdot\text{Attention}({Q},{K}',{V}'),
\end{equation}
where $K' = c_{i}{W}'_{k}$,$V' = c_{i}{W}'_{v}$, $c_{i}$ represents the image features extracted from the CLIP image encoder, and $Z^{new}$ denotes the new latent representation obtained by conditioning on both image and text inputs. 

\subsection{Framework of Character-Adapter}

In this section, we introduce Character-Adapter, a novel framework designed for high-fidelity consistent character generation. We introduce prompt-guided segmentation to ensure fine-grained regional features of reference characters
and dynamic region-level adapters to mitigate concept confusion. As illustrated in Fig. \ref{fig:overview1}, we first segment a reference character into several parts using prompt-guided segmentation. Subsequently, We obtain the attention maps of the target image layout using the same approach. Finally, we utilize dynamic region-level adapters to achieve detailed and consistent character generation.

\subsubsection{Prompt-guided Segmentation}
\label{Prompt-guided segmentation}
Directly passing the entire reference image into the image encoder can lead to inadequate extraction of detailed image features related to the given character. It potentially causes identity degradation and compromises detailed preservation. To address this, one approach is to divide the entire reference image into essential parts, enabling the image encoder to focus on specific regions of an image for comprehensive feature extraction. We propose a prompt-guided segmentation module, localizing essential image regions based on text prompts, thereby facilitating adequate image feature extraction.

Considering the fundamental aspects of a character, we categorize it into three parts: the face, upper body, and lower body. When given a user-input prompt $P$, for instance, \emph{``a boy standing in a library, wearing green jacket and blue pants''}, we complete the prompt with detailed region descriptions:
\begin{equation}
    \label{eq_prompt}
    P = \{P_{m}\} = \{w_{i}\}, m=F,U,L \quad i=1,...,N,
\end{equation}
where $N$ refers to the number of words in the prompts, $P_F$ indicates the prompt for the face region (\emph{``a boy''}), $P_U$ specifies the prompt for the upper body (\emph{``green jacket''}), and $P_L$ relates to the prompt for the lower body (\emph{``blue pants''}).  $W:$ represents the sequence of words in the constructed prompt $P$.

Previous works prove the cross-attention maps in the diffusion model contains semantic information of a latent image. Character-Adapter starts with this well-explored finding, seeking to use cross-attention maps between specific text prompts and the latent image to localize the three part of a character. Specifically, given a prompt $P$, the attention map between each word $w_{i}$ and the image latent of $k$-th layer of the UNet can be defined as the following:
\begin{equation}
\begin{split}
    \label{eq_layout_attn}
    A_{i, t}^{(k)} &= F(w_i, z_t^{k}) \\
    &= softmax((W_q^{(k)} \cdot z_t^{k})(W_k^{k} \cdot \mathcal{E}(w_i))),
\end{split}
\end{equation}
where $A_{i, t}^{(k)}$ denotes the attention map between the word $w_{i}$ and the image latent $z_{t}$ in the $k$-th layer of UNet, $W_q$ and $W_k$ are projection matrices, and $\mathcal{E}$ corresponds to a language model that encodes text prompts into text embeddings. To obtain precise attention maps, we integrate the attention maps of all $K$ layers in the UNet through interpolation to obtain the final attention map $A_{i, t}$ between $w_i$ and the image latent $z_t$ as the following:
\begin{equation}
    A_{i, t} = \sum_{k=1}^{K}A_{i, t}^{(k)},
    \label{eq:wordattentionmap}
\end{equation}
where $K$ denotes the total number of layers in the UNet.

A region prompt, denoted as $P_{m}$ in Eq. \ref{eq_prompt} contains several words (\emph{``green jacket''}). Consequently, we aggregate the attention maps associated with these words, denoted by $A_{i, t}$ in Eq. \ref{eq:wordattentionmap}, to obtain the cross-attention map between each region prompt and the latent image:
\begin{equation}
\label{eq_sim}
    A_{r, t} =  \mathop{\max}\limits_{r_{begin} < i < r_{end}} A_{i, t}, 
\end{equation}
where $r_{begin}$ and $r_{end}$ represent the word indices corresponding to the beginning and end of the region prompt, respectively. Therefore, we utilize the attention maps $A_{r, t}$ to partition an image into each region part $I_{P}$ mentioned in Eq. \ref{eq_prompt}:
\begin{equation}
\label{eq_region1}
y_{1} = \mathop{\arg\min}\limits_{y} (A_{r}(x,y) > \gamma_1) , \quad x_{begin}^{r} = x_{1} ,
\end{equation}
\begin{equation}
\label{eq_region2}
y_{2} = \mathop{\arg\max}\limits_{y} (A_{r}(x,y) > \gamma_1) , \quad x_{end}^{r} = x_{2},
\end{equation}
\begin{equation}
\label{eq_region3}
I_{P} = [x_{1},x_{2},y_{1},y_{2}]
\end{equation}
where $x$,$y$ represents the width and the height of the original image, $\gamma_1$ is  set to 0.8 in our paper.

As illustrated in Stage 1 in Fig. \ref{fig:overview1}, a reference image $I_{R}$ is segmented into three main parts by using Prompt-guided Segmentation. Specifically, we add noise to the latent representation $z_0^{R}$ of the reference image $I_{R}$ in $t$ steps as the following:
\begin{equation}
z_t^{R}=\sqrt{\overline{\alpha}_t}z_0^{R}+\sqrt{1-\overline{\alpha}t}\epsilon_t .
\end{equation}
Following Eq. \ref{eq_prompt}-\ref{eq_region3}, a reference image $I_{R}$ is ultimately split into three parts:  $I_{P_F}$,$I_{P_U}$,$I_{P_L}$.

\subsubsection{Dynamic Region-level Adapters}

\label{Dynamic region-level adapters}
Given that cross-attention layers are effective in controlling the appearance of the generated image, IP-Adapter proposes an image prompt adapter to inject the feature of a reference image into the diffusion model. However, when multiple image conditions are added to the diffusion model, it becomes unable to generate an image that preserves each of the conditional features, leading to concept fusion. To mitigate this limitation, we propose a dynamic region-level adapters module, using specific regions' layout to alleviate concept fusion during the inference process.

Upon obtaining each region of the reference image $I_{P_{F}}$,$I_{P_{U}}$,$I_{P_{L}}$ mentioned in Eq. \ref{eq_prompt}, we mainly inject them into diffusion model using a \textit{region-level adapters} and \textit{Attention Dynamic Fusion} module.
\begin{algorithm}[!t]
    \small
    \caption{\label{alg}Character-Adapter for single and multi-characters consistency generation}
    \begin{algorithmic}[1]
        \REQUIRE{Prompt $P$, $N$ reference character images $I_i, i \in [1, N] $, a diffusion network $D_\theta$ }
       \STATE  get the regions of the reference image via Eq. \ref{eq_region3}
        \FOR{i in [1, N]}
            
             \STATE use prompt $P_{m}$ to obtain attention map $A_{i, t}^{k}$ via Eq. \ref{eq_layout_attn} and correlation map $A_{r, t}$ via Eq. \ref{eq_sim} 
            \STATE get region adapter outputs $I_{P}$ via Eq. \ref{eq_mask_adapter1}
        \ENDFOR
        \STATE fuse the adapters into U-Net to get the hidden state via Eq. \ref{eq:ipadapter} 
        \ENSURE{target image $T_I=D_\theta(A_t)$}
    \end{algorithmic}
\end{algorithm}
\textbf{Region-level Adapters.} 
Region-level Adapters consist of three individual IP-Adapters, with each IP-Adapter being responsible for injecting features of one region of the reference image. Obtaining the specific layout regions corresponding to each region of the reference image is crucial. In our task, we generate an image with a given prompt and a reference image. During the denoising process, the cross-attention maps between the region prompts and the image latent can be obtained by using Prompt-guided Segmentation mentioned in Eq.\ref{eq_prompt}-\ref{eq_sim}. These attention maps can be served as the masks of three regions of the layout image, as illustrated in Stage 2 in Fig. \ref{fig:overview1}. 
The masks at $(x, y)$ of each region can be obtained as following:
\begin{equation}
\label{eq_mask}
\begin{split}
    M_{r,t}(x, y)&=Binary(A_{r,t}(x,y)) \\
    &=\left\{\begin{array}{l} 
    1 \quad \quad A_{r,t}(x,y) > \gamma_2, \\
    0 \quad \quad A_{r,t}(x,y) <= \gamma_2.\\ 
    \end{array}  x \in [1, w], y \in [1, h] \right.,
\end{split}
\end{equation}
where $A_r(x,y)$ is the attention map obtained by Eq. \ref{eq_sim}. $\gamma_2$ is the threshold, we set it as 0.8 in our paper.

Since these adapters are tasked with handling distinct regions during image generation, such as using the face adapter to influence the character's appearance and hairstyle, a logical method involves merging the various adapters through mask guidance. The current masked attention map between a region prompt $P_{m}$ a latent image $z_{t}$ is as follows:
\begin{equation}
    A_{r, t}^{R} =  \sum\limits_{x=0}^{w}\sum\limits_{y=0}^{h}(A_{r,t}(x,y) \cdot M_r(x, y)) ,
\label{eq_mask_adapter}
\end{equation}
where $A_{r, t}(x,y)$ can be obtained from Eq. \ref{eq_sim} using $P_{m}$. $w$,$h$ refers to the resolution of the latent image $z_{t}$.
Subsequently, the current masked attention map is sent into the s elf-attention module. The latent image after the self-attention module can be calculated as:
\begin{equation}
    e_{P_{m}}^{i} = A_{r, t}^{R} W_{v}^{i} \mathcal{E}(I_{P_{m}})
\label{embeddingone}
\end{equation}
where $\mathcal{E}$ denotes the image encoder.
The output latent of the $i$-th layer in UNet by using region-level adapters is summed as following:
\begin{equation}
    {z'}_{t}^{i} = z_{t}^{i}+e_{P_{F}}^{i}+e_{P_{U}}^{i}+e_{P_{L}}^{i},
\label{embedidng}
\end{equation}
where $e_{P_{F}}$,$e_{P_{U}}$ and $e_{P_{L}}$ denotes the output of each region-level adapter of face region, upper region and lower region, which can be obtained by Eq. \ref{embeddingone}.

\textbf{Attention Dynamic Fusion.} Mask-based regional-level adapters face the challenge of obtaining region masks that facilitate generation adhering to the layout while remaining faithful to the reference image. To address this, we propose an attention dynamic fusion module to further improve the quality of the generated image. Specifically, we characterize the masks as soft-label masks by removing $M_{r}(x, y)$ mentioned in Eq. \ref{eq_mask} and integrate them into region-level adapters to produce coherent image generation. Thus, The output of the region-level adapters can be rewritten as following: 
\begin{equation}
\begin{split}
    A_{r, t}^{R} &= \sum\limits_{x=0}^{w}\sum\limits_{y=0}^{h}A_{r,t}(x,y) \\
\end{split}
\label{eq_mask_adapter1}
\end{equation}
Then $A_{r, t}^{R}$ is passed into Eq. \ref{embeddingone}-\ref{embedidng} to obtain the hidden state of the $i$-th layer in UNet using dynamic region-level adapter.


\subsubsection{Multi-character Generation}
In addition to achieving remarkable performance in terms of single-character consistency, Character-Adapter also demonstrates remarkable performance for multi-character generation. Similar to single-character generation, we utilize prompt-guided segmentation to automatically localize multiple reference characters. The prompt for $J$ character is noted as:
\begin{equation}
P = \{P_{m}^{j}\}, m=F,U,L \quad j=1,...,J,
\label{eq_prompt_n}
\end{equation}
The process of multi-character consistency generation is described in Algorithm \ref{alg}.

\section{Experiments}

\subsection{Experimental setup}

\label{Experimental setup}
\textbf{Evaluation Datasets.} 
We collect a dataset comprising 50 distinct entities, which includes 25 real-world characters and 25 anime characters for single-character evaluation. For multi-character evaluation, we randomly combine two characters from real-world characters and anime characters, creating a total of 50 samples. All user-given texts are obtained from MSCOCO \cite{lin2014microsoft} datasets.

\textbf{Implementation Details.}
 We conduct all experiments on SD v1.5 \cite{rombach2022high} with classifier-free guidance as 7 and 20-timestep Euler A sampling. The mask threshold $\gamma_1$ and $\gamma_2$ are all set as 0.8. We use three identity IP-Adapter-Plus in dynamic region-level adapters based on SD1.5. We conducted our experiments using NVIDIA A30 GPUs, more details are provided in Appendix.

\subsection{Quantitative Comparison}
\label{Quantitative comparison}

\begin{table*}[!t]
\centering
\small
\setlength{\tabcolsep}{0.8mm}{
\begin{tabular}{cc|ccccccccccccc}
\toprule
\multirow{2}{*}{Methods} & \multirow{2}{*}{Models} & \multirow{2}{*}{Time(s)} & \multicolumn{2}{c}{CLIP-T\textbf{(\%)}$\uparrow$} &  & \multicolumn{2}{c}{CLIP-I\textbf{(\%)}$\uparrow$} & & \multicolumn{2}{c}{DINO-I\textbf{(\%)}$\uparrow$}     \\
\cline{4-5} \cline{7-8} \cline{10-11}
& & & Single & Multi& & Single & Multi & & Single & Multi &\\
\midrule
\multirow{6}{*}{Finetune-based}                 
                               & Textual Inversion \cite{gal2022textual}      &220  & 26.2             &   24.0       &&  82.8              &   82.1      &&  52.3     & 42.8         \\
                               & LoRA \cite{hu2021lora}       &1050         &   27.3              &    25.4      &&  83.5             &   83.2       &&  61.2    &  62.4        \\
                               & Custom Diffusion \cite{kumari2023multi}  &510    &    28.6            &   24.8       &&   83.9             &   83.6       && 67.8     &   67.2       \\
                               & Mix-of-Show \cite{gu2024mix}     & 1200      &   \underline{29.9}                &  27.4         &&  84.2             &    \underline{84.1}    &&  \underline{68.2}   &   65.9       \\
                               & OMG \cite{kong2024omg}          &1200        &   29.7                 &   \underline{28.6}       &&  \underline{84.6}              &    83.6      &&  67.1 &   \underline{67.8}       \\ \midrule
\multirow{7}{*}{Training-free} & Reference only\cite{ref:reference_only}     &7.6    & 26.8          &  22.8        && 78.4          &   73.6        &&  42.1   &   40.8       \\
                               & IP-Adapter \cite{ye2023ip}      &5.2      & 29.8          &  26.2        && 83.6           &   82.4       &&  59.8   &   58.0       \\
                               & InstantID \cite{wang2024instantid}       &5.8   & 27.4          &  22.1        && 75.8          &   74.2      &&  46.7     & 45.9         \\
                               & FastComposer \cite{xiao2023fastcomposer}    &7.8      & 26.4          &  21.1        && 72.4          &   67.2      &&  42.7     & 41.9         \\
                               & T2I-Adapter \cite{mou2023t2iadapter}     &7.6       & 24.2          &  21.8        && 63.8          &   62.3      &&  43.4     & 40.9         \\
                               & BLIP-Diffusion \cite{li2024blip}     &4.5       &  29.7               &   27.4       &&  84.0             &  82.7         && 60.2    &    53.2      \\
                               & \textbf{Character-Adapter (Ours)}           &7.2       & \textbf{30.4}  &  \textbf{30.2}  && \textbf{84.8} & \textbf{84.6}&& \textbf{68.1}& \textbf{67.8}     \\
             
\bottomrule
\end{tabular}
}
\caption{\label{tab:Quantitative}Quantitative results (\%) of Character-Adapter with other methods. Evaluations are conducted for both single and multiple character generation. The best results of training-free methods are highlighted in bold. The best results of finetune-based methods are highlighted in underline.}
\end{table*}
We conduct experiments on  Character-Adapter and previous SOTA works, quantitative results are illustrated in Tab. \ref{tab:Quantitative}. state-of-the-art performance in zero-shot character consistency generation, with an improvement of 24.8\% on the CLIP-I\cite{clip} and DINO \cite{dino} scores compared to previous state-of-the-art methods in both single and multi character generation. In comparison with fine-tuning-based approaches, Character-Adapter can achieve comparable character consistency while offering over 70 times improvement in computational efficiency, as it requires no additional training. Moreover, Character-Adapter achieves the best performance in terms of text-image alignment with an improvement of 3.5\%.


\subsection{Qualitative Comparison} \label{Qualitative comparison}

\begin{figure*} [!t]
\centering
\includegraphics[width=0.9\linewidth]{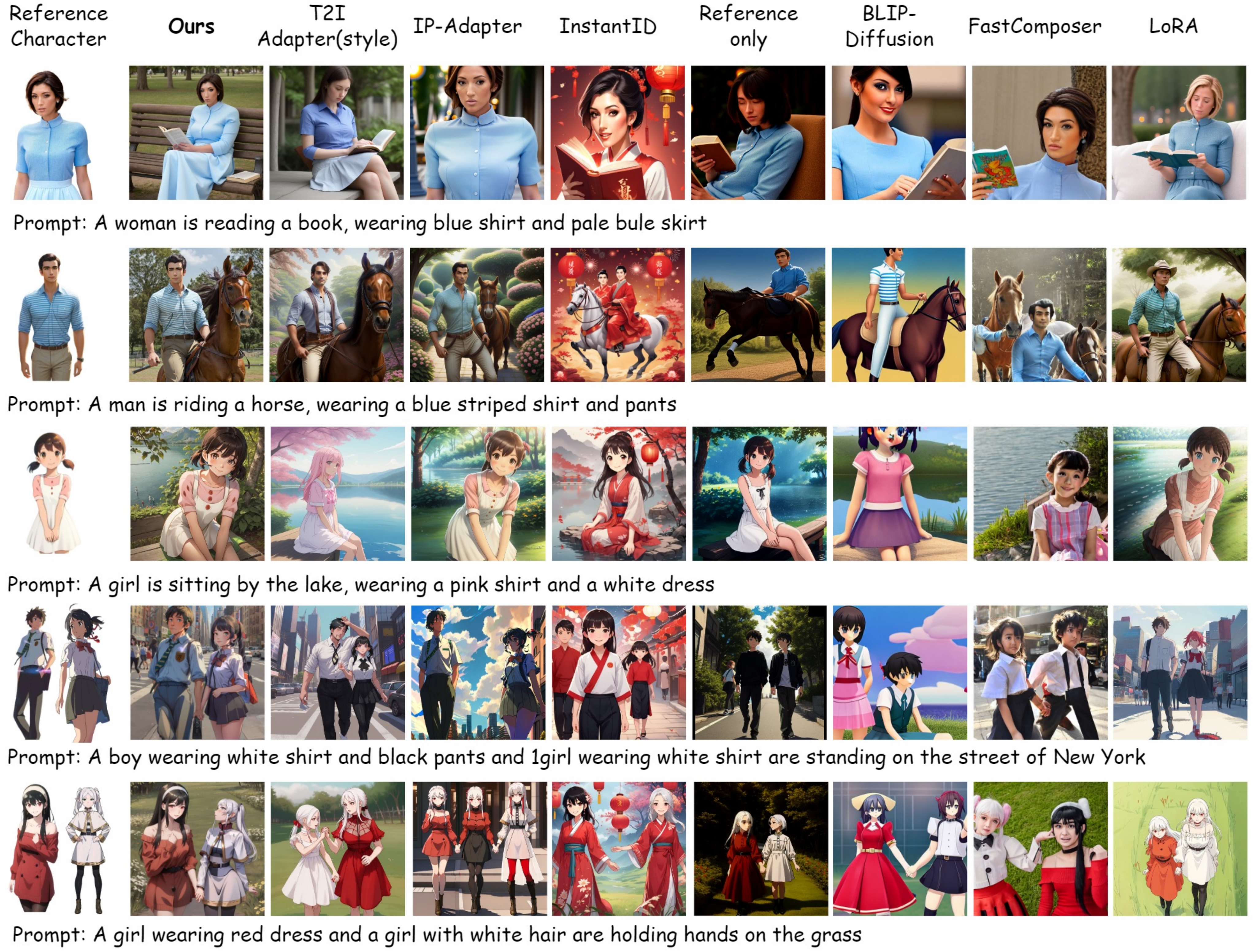} 
\caption{\label{fig:case} Visual comparison (\%) of Character-Adapter against other subject-driven methods. Our approach ensures high-fidelity consistency, while maintaining text-image alignment. } 
\end{figure*}

As shown in the multi-character generation example (Fig. \ref{fig:case} 6th row), methods like T2I-Adapter, InstantID, Reference-only, and FastComposer fail to preserve the reference characters' attire. IP-Adapter encounters concept fusion and detail omission issues, attributable to using an image-level adapter, leading to inadequate feature extraction and representation. While fine-tuning-based models like LoRA align with text prompts, they lack details of reference characters and cause concept fusion in multi-character generation. In contrast, Character-Adapter generates high-fidelity multi-character images with intricate details while preserving text-image alignment. Overall, the qualitative results substantiate Character-Adapter's significant improvements in high-fidelity character generation. Additional results are provided in Appendix. 

\subsection{Ablation Study} \label{ablation}

\begin{figure} [!t]
\label{fig:abl}
\centering
\includegraphics[width=0.8\linewidth]{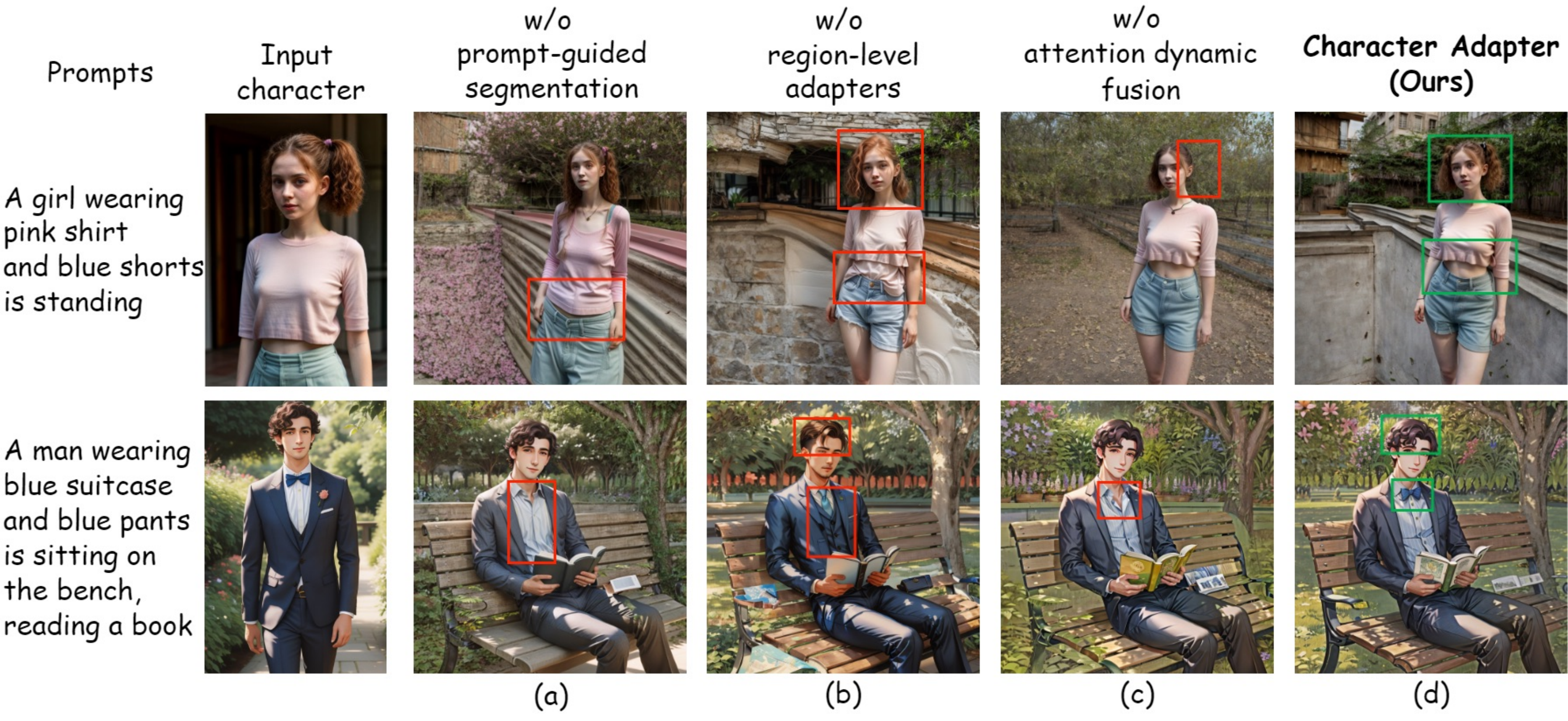} 
\vspace{-5pt}
\caption{\label{fig:ablation} Visualization of ablation study, each component is removed individually to prove its efficiency, (d) represents the results obtained with the whole Character-Adapter. } 
\vspace{-4pt}
\end{figure}

\begin{table}[!t]
    \centering
    \small
    \setlength{\tabcolsep}{1.2mm}
    \begin{tabular}{c|ccc}
    \toprule
    \makebox[0.1\textwidth][c]{\textbf{Methods}} &
                                     \makebox[0.2\textwidth][c]{CLIP-T$\uparrow$} & \makebox[0.2\textwidth][c]{CLIP-I$\uparrow$} & 
                                     \makebox[0.2\textwidth][c]{DINO-I$\uparrow$}            \\

    \midrule
    \ w/o \{Prompt-guided segmentation\} &  26.9 & 77.9 & 47.2\\
    \ w/o \{Region-level adapter\}& 28.4 & 79.8 & 53.9 \\
    \ w/o \{Attention dynamic fusion\} & 27.4 & 78.3 & 62.2  \\
    \textbf{Character-Adapter (full)} & \textbf{30.8} & \textbf{86.9} & \textbf{68.3} \\
    \bottomrule
    \end{tabular}
    \caption{\label{tab:abl}Quantitative ablation result (\%) of Character-Adapter. Each component is individually removed to evaluate its necessity and contribution to the overall performance.}
    \vspace{-4pt}
\end{table}

\begin{table}[!t]

   \centering
  \setlength{\tabcolsep}{1mm}{
    \begin{tabular}{cccccc}
     \toprule
      \multicolumn{1}{c}{\multirow{2}[4]{*}{\thead{Models\\ (Ours vs. *)}}} &  \multicolumn{2}{c}{Text Alignments} & & \multicolumn{2}{c}{Character Consistency}  \\
 \cmidrule{2-3} \cmidrule{5-6}
       & Win(\%)  & Lose(\%) & & Win(\%) & Lose(\%)  \\
    \midrule
      LoRA & 92.4 & 7.6& & 90.6& 9.4 \\
      \midrule
      IP-Adapter  & 87.5 & 12.5& & 70.8 & 29.2 \\
        T2I-Adapter & 90.2 & 9.8& & 91.7& 8.3 \\
       InstantID & 94.2 & 5.8& & 100& 0.0 \\
     Reference-only & 100 & 0.0& & 99.2& 0.8 \\
      BLIP-Diffusion & 100 & 0.0& & 100& 0.0 \\
      FastComposer & 98.2 & 1.8& & 100& 0.0 \\
    \bottomrule
     \end{tabular}%
     }

     \caption{\label{tab:user_study} User study results(\%) illustrating the comparison between Character-Adapter and SOTA methods.}
   \vspace{-8pt}
 \end{table}
\textbf{Prompt-guided Segmentation.} 
We conduct an experiment where the character is equally segmented into three parts, and compare the results to verify the importance of prompt-guided segmentation. Fig. \ref{fig:ablation} (a) exhibits a significant imbalance in the proportions of the characters, in contrast to Fig. \ref{fig:ablation} (d). The results indicate that prompt-guided segmentation facilitates comprehensive image feature extraction and provides fine-grained guidance for character generation. 

\textbf{Region-level Adapters.} 
The experimental results in Table \ref{tab:abl} show a significant drop in the CLIP-I score when removing the region-level adapters. A comparison between Fig. \ref{fig:ablation} (b) and Fig. \ref{fig:ablation} (d) demonstrates that the proposed region-level adapters effectively mitigate content fusion issues, thereby improving character consistency. According to Tab. \ref{tab:region-selection}, we can observe that as the number of regions increased, our model achieved better performance across all metrics. However, when the number of regions reached 3 and 4, the performance improvements became marginal with no significant growth. Consequently, we opted for 3 regions as the optimal configuration.

\textbf{Attention Dynamic Fusion.} Fig.\ref{fig:ablation} (c) showcases the result without using  attention dynamic fusion. The visualization in Fig. \ref{fig:ablation} (b) reveals a dramatic change in the character's hairstyle compared to Fig.\ref{fig:ablation} (d). This is because the solid mask only preserves details of certain regions of the reference character and is influenced by the layout image. The result in Tab. \ref{tab:abl} also corroborates the effectiveness of the proposed attention dynamic fusion module.

\begin{table}[!t]
   \centering
   \small
 \setlength{\tabcolsep}{0.6mm}{
    \begin{tabular}{cccccccccc}
     \toprule
      \multicolumn{1}{c}{\multirow{2}[4]{*}{\thead{Methods}}} &  \multicolumn{2}{c}{CLIP-T(\%)} & & \multicolumn{2}{c}{CLIP-I(\%)} & & \multicolumn{2}{c}{DINO-I(\%)} \\
 \cmidrule{2-3} \cmidrule{5-6} \cmidrule{8-9}
       & Single & Multi & & Single& Multi && Single & Multi  \\
    \midrule
 
     1-region(IPA)  &29.8 &26.2	& & 83.6 &82.4 &&59.8 &58.0 \\
       2-region &29.4 & 28.3 & & 83.9& 82.8 && 60.2 & 58.2 \\
       \textbf{3-region(Ours)} & \textbf{30.4} & \textbf{30.2}& & \textbf{84.8}& 84.6 && \textbf{68.1} &\textbf{ 67.8} \\
     4-region & 30.2 & \textbf{30.2}& & 84.2&\textbf{ 84.8} && 67.4 & 66.9  \\
    
    \bottomrule
     \end{tabular}
     }
     \caption{Quantitative results (\%) of Character-Adapter with different prompt-guided regions, compared with IP-Adapter(IPA). The best results are highlighted in bold.}
     \label{tab:region-selection}
   \vspace{-10pt}
\end{table}

\begin{table}[!t]
    \centering
    \small
    \begin{tabular}{ccccc}
    \toprule
       Method  & 1-char & 2-char & 3-char &4-char \\
       \midrule
       Time (s)  &  7.2 & 8.78 & 9.32 & 10.4 \\
      VRAM(MB)   &  5940 & 8116 & 10536 & 12520\\
      \bottomrule
    \end{tabular}
    \caption{Inference time and computational efficiency of Character-Adapter in generating varying numbers of characters (n-char) compared with IP-Adapter(IPA).}
    \label{tab:compare_ipa}
\end{table}

\subsection{Computational Study}
Furthermore, we conduct experiments on the inference time and computing resources of generation 1 to 4 characters with resolution of 768 $\times$ 768. As illustrated in Tab. \ref{tab:compare_ipa}, increasing the number of characters from 1 to 2 results in an average increase in inference time of just 1.07 seconds, which represents a 14.9\% increase compared to the inference time for a single character. As the number of characters increases, the computational load and memory usage do not exhibit linear growth. This demonstrates the superior engineering efficiency of Character-Adapter. And the A30 GPU is sufficient for running our model, which demonstrates the superior engineering efficiency of Character-Adapter.

\subsection{User Study} 
\label{user study}
We conducted a user study with 20 experts to evaluate Character-Adapter against previous methods. Each expert assessed 50 pairwise comparisons. As shown in Tab. \ref{tab:user_study}, the results indicate that Character-Adapter achieves a win rate significantly exceeding 50\% for both text alignment and character consistency.

\begin{figure}[!t]
\centering
\includegraphics[width=1\linewidth]{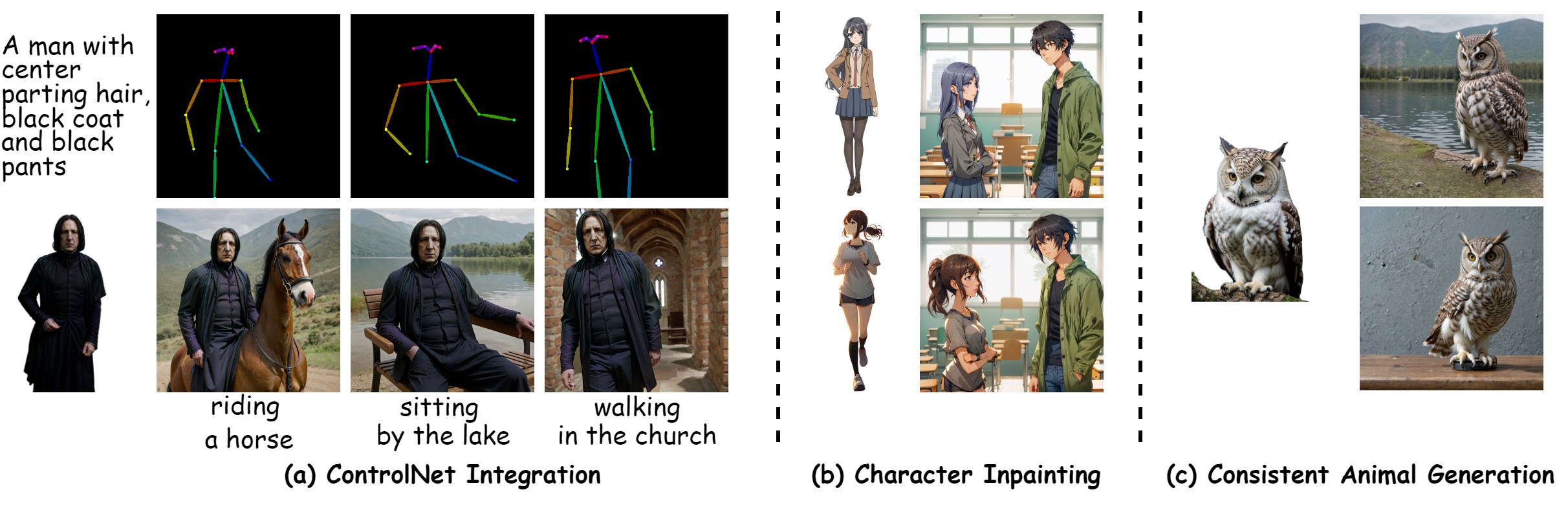} 
\vspace{-10pt}
\caption{\label{fig:pose_control} Visualization of Character-Adapter's versatility and compatibility. (a) Combination with Pose Control. (b) Inpainting with a reference image. (c) Generation with animals (other types).} 
\end{figure}
\subsection{Extended Application}

\textbf{Additional controls.} Fig. \ref{fig:pose_control} demonstrates that Character-Adapter can be successfully integrated with ControlNet, enabling pose conditioning to guide the consistent character generation.

\textbf{Inpainting-based personalization.} Our method can also be combined with inpainting to seamlessly insert consistent characters into any fixed scene, which holds great potential for generating comics and graphic novels. Fig. \ref{fig:pose_control} shows that integrated with inpainting, our proposed Character-Adapter can replace a selected region with specific characters, while maintaining a fixed background naturally.

\textbf{Consistent Generation of Non-human Characters.}
As Character-Adapter supports character consistency, it has the potential to extend the ability to animals and objects. In Fig. \ref{fig:pose_control}, we have demonstrated that Character-Adapter can achieve excellent results in the task of animal consistency.

\section{Conclusion}

In this work, we investigate the limitations of existing consistent character generation methods and propose Character-Adapter, which enables the synthesis of high-fidelity images that preserve the intricate details and identities of reference characters. The key innovations in Character-Adapter are the prompt-guided segmentation module, which enables fine-grained extraction of regional features from reference characters, and the dynamic region-level adapters module, which mitigates concept fusion issues. Notably, Character-Adapter can be seamlessly integrated into any backbone model and compatible with other editing tools for both single and multiple character generation.


\newpage
{
\small
\bibliographystyle{unsrt}
\bibliography{main}
}
\newpage
\appendix
\section{Appendix}
\label{appendix}

We have provided supplementary details regarding our Character-Adapter in this section. 

\subsection{Implementation details}
\label{implementaion}

\subsubsection{Inference setup}
\label{inference setup}
We employ Realistic Vision V4\footnote{https://huggingface.co/SG161222/Realistic\_Vision\_V4.0\_noVAE} to generate photo-realistic human portraits and animals, and using Animesh\footnote{https://huggingface.co/redstonehero/animesh\_prunedv21} for anime character generation. All comparison models utilize 20-step Euler A sampling, and the classifier-free guidance is set to be 7.0. The corresponding resolution of inference image is set to 768 x 768. We implement experiments using an A30 GPU. 

\subsubsection{Evaluation metrics}
\label{metrics}
We first employ CLIP ViT-L/14\footnote{https://huggingface.co/openai/clip-vit-large-patch14} to evaluate the similarity between the generated images and the given text prompts (\textbf{CLIP-T}). Subsequently, we utilize the image encoder of the CLIP model to evaluate the correlation between the generated consistent images and the reference images (\textbf{CLIP-I}). Additionally, we further employ the DINO score \cite{liu2023grounding} to evaluate image alignment, as DINO is better suited for subject representation (\textbf{DINO-I}). Finally, we conduct human evaluations to further evaluate the performance of different approaches in terms of text and image alignment.

\subsection{Segmentation Evaluations}
In this subsection, we elucidate the differences between our proposed prompt-guided segmentation and Grounded Segment Anything(Grounded-SAM)\cite{ren2024grounded}. Additionally, we explain our rationale for not selecting existing open-world semantic segmentation models, such as Grounded-SAM.

\begin{figure}
    \centering
    \includegraphics[width=1\linewidth]{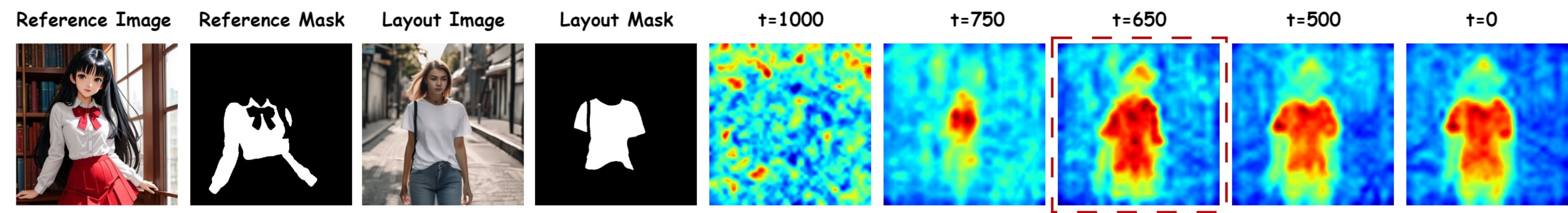}
 \caption{Visualization of attention maps between the upper body region prompt and image latent. As
the timestep increases, the attention map more accurately reflects the area of the character clothing.}
    \label{fig:heatmap}
\end{figure}

\begin{figure}
    \centering
    \includegraphics[width=1\linewidth]{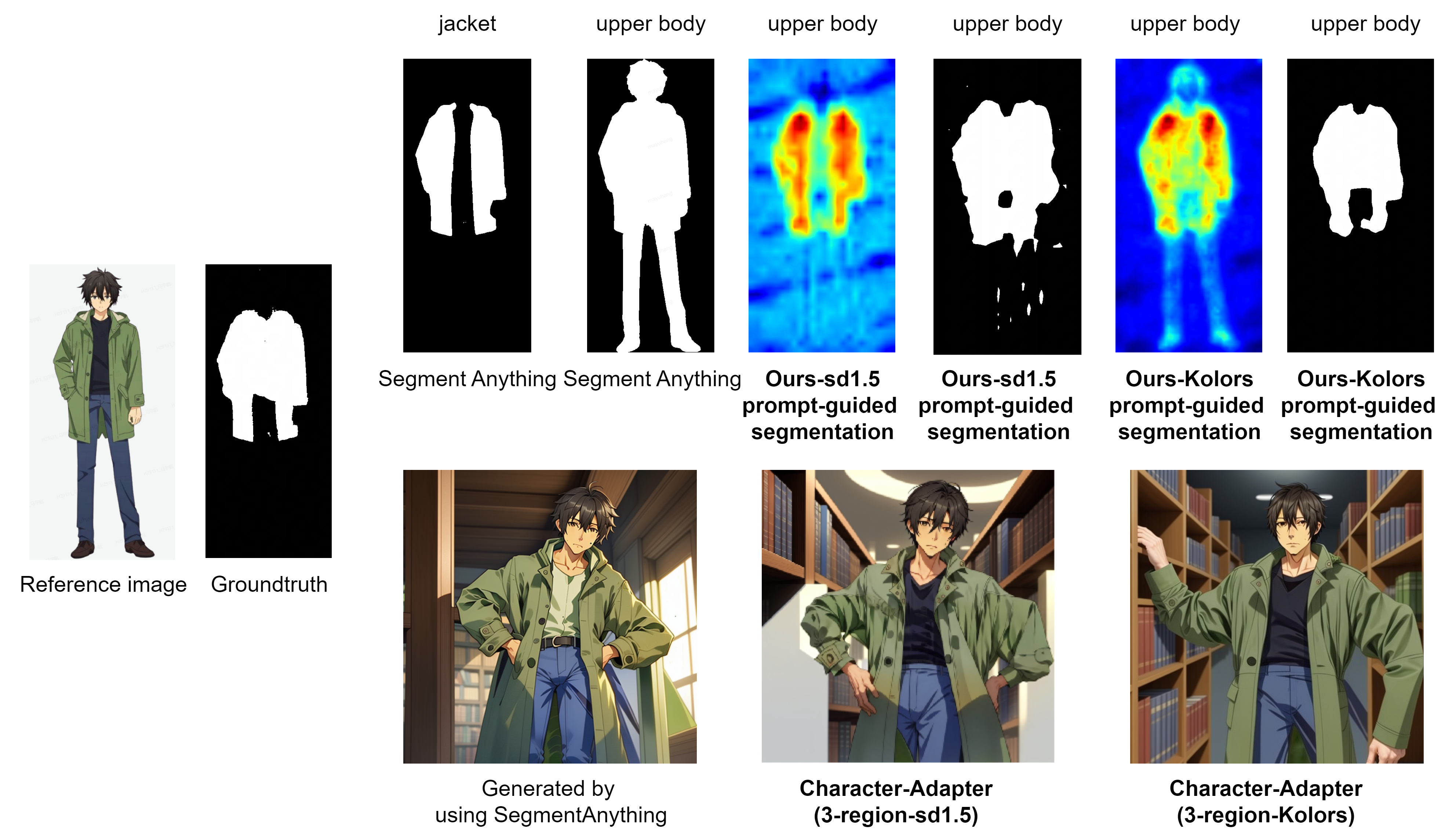}
 \caption{Visual comparison of reference image masks obtaining by our proposed prompt-guided segmentation and Segment Anything}
    \label{fig:segmentation}
\end{figure}

Segmenting from a pre-trained model exhibits a noticeable disparity between our layout image and the reference image as illustrated in Fig. \ref{fig:heatmap}. However, when using Grounded-SAM, the accuracy is so high that we are unable to obtain the entire upper body. Our objective is to ensure minimal overlap between the masks of the three parts we aim to extract, thereby maintaining the integrity and continuity of image features. 
We conducted an experiment using a reference image for demonstration purposes, which is provided in Fig. \ref{fig:segmentation}. When attempting to segment the upper body using Grounded-SAM with the prompt “jacket”, we only capture the jacket region, necessitating the use of a threshold to acquire the complete upper body of the reference image. Conversely, using “upper body” as a prompt results in the model returning the entire body. Prompt-guided segmentation, while providing a rough masked region, enables us to obtain a relatively complete and expected mask. 
In addition to conducting experiments on SD1.5\cite{rombach2021highresolution}, we also validated the effectiveness of our prompt-guided segmentation approach on Kolors \cite{kolors}. This confirms that our proposed plug-and-play method is applicable to any text-image cross-attention-based model. As illustrated in the figure, the character's distinctive features are effectively preserved. This further demonstrates that a well-designed model with strong semantic understanding contributes to generating more precise cross-attention maps, ultimately leading to results with higher consistency.
Furthermore, as illustrated in Tab. \ref{tab:sam}, the use of Grounded-SAM significantly increases GPU memory usage, approximately by 10 GB, leading to a substantial increase in computing resource consumption. Consequently, we employ prompt-guided segmentation to partition characters into multiple distinct regions.

\begin{table}[t]
    \centering
    \setlength{\tabcolsep}{0.5mm}
    \scalebox{1}{
    \begin{tabular}{ccccccc}
    \toprule
        Method & Mask IoU(\%) &CLIP-T(\%) &CLIP-I(\%) &DINO-I(\%)  &VRAM(MB)  & Time(s)  \\
        \midrule
        Grounded-SAM & 63.5 & 30.4 & 84.5 & 67.9 & 8549 &2.2 \\
       Prompt-guided Segmentation  & 62.5 & 30.4 & 84.8&68.1 &5592  & 1.17 \\
       \bottomrule
    \end{tabular}}
    \caption{Quantitative results of Prompt-guided Segmentation and Grounded-SAM.}
    \label{tab:sam}
\end{table}



\subsection{Hyperparameter Setup}
We performed a comprehensive hyperparameter ablation study using the evaluation datasets described in our paper. By systematically fixing one parameter while adjusting the other, we were able to determine optimal values for both. The results of this analysis are presented in Tab. \ref{tab:threshold}.

\begin{table}[t]
    \centering
        
        \scalebox{1}{
    \begin{tabular}{c|c|ccc}
     \toprule
        $\gamma_{1}$ & $\gamma_{2}$ & CLIP-T\textbf{(\%)}$\uparrow$ & CLIP-I\textbf{(\%)}$\uparrow$  & DINO-I\textbf{(\%)}$\uparrow$\\
        \midrule
        0.8 & 0.7  & 25.5 & 79.25 & 67.4 \\
        0.8 & 0.8 & \textbf{30.4} & \textbf{84.8}  & 68.1\\
       0.8  & 0.9 & \underline{29.5} & \underline{83.4} & \textbf{ 71.2} \\
        0.7 & 0.8 & 26.2 & 74.9 & 67.87\\
        0.9 & 0.8 & 27.8 & 78.76 & \underline{68.84}\\
        \bottomrule
    \end{tabular}
}
    
    \caption{\label{tab:threshold}Quantitative results (\%) of Character-Adapter under different thresholds.}
\end{table}

\subsection{Limitations and discussion}
\label{limitaion}

While our method provides a plug-and-play framework for generating consistent and detailed characters with high robustness, several limitations warrant consideration. Firstly, in scenarios involving extremely complex clothing patterns, our model may not fully preserve the original details. Secondly, due to the inherent limitations in the semantic capabilities of  Stable Diffusion models, there exists a possibility of inaccurate target localization in the attention maps, leading to misalignment of image details. We leave the exploration of these limitations as future work.

\subsection{Future work}
\label{future}
For future work, an intriguing direction would be to investigate methods for obtaining more accurate attention maps for different image regions based on prompts, thereby mitigating semantic confusion. This could be achieved by enhancing the semantic understanding of the diffusion model. Furthermore, as Character-Adapter supports identity preservation for both facial and attire features, it holds promise for applications in narrative storytelling and video generation. 

\subsection{Societal impacts}
\label{societal}
The proposed method aims to provide an effective and flexible tool for high-fidelity customized character image generation. However, there exists a potential risk of misuse, wherein individuals may generate fake celebrity images, thereby misleading the public. This concern is not unique to our approach but rather a common consideration among all subject-driven image generation methods. One potential solution is to employ a safety checker akin to NSFW filter,\footnote{https://huggingface.co/runwayml/stable-diffusion-v1-5} which is a classification module that estimates whether generated images could be considered offensive or harmful, thereby preventing the generation of controversial content and the abuse of celebrity images. Such measures would mitigate the potential misuse of our method while preserving its intended functionality.

\subsection{Additional results}
\label{additionalresults}

\begin{figure*} [thb]
\centering
\includegraphics[width=0.95\linewidth]{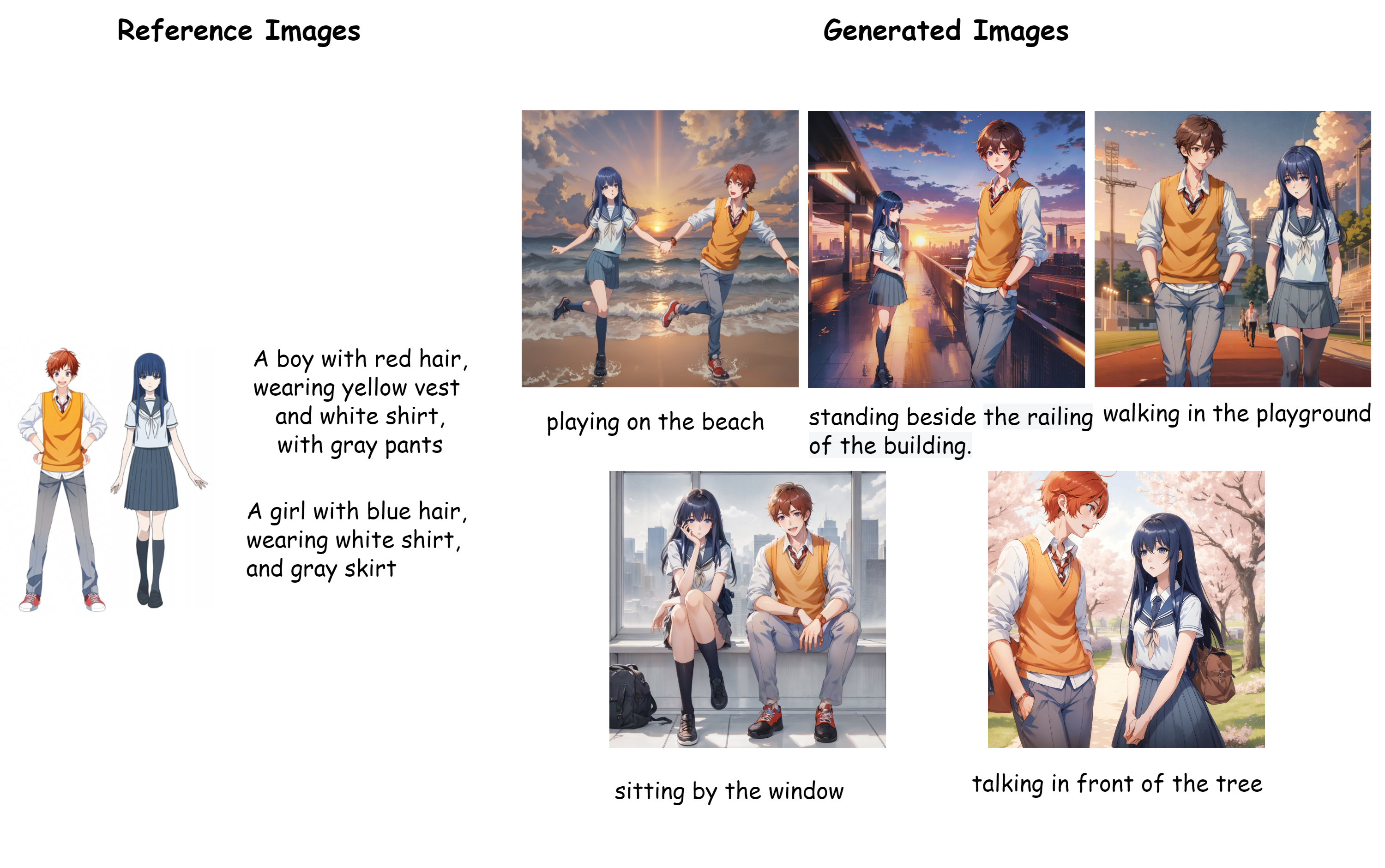} 
\caption{\label{fig:additional_result2} Additional qualitative results of multi characters. }
\end{figure*}

\begin{figure*} [thb]
\centering
\includegraphics[width=1\linewidth]{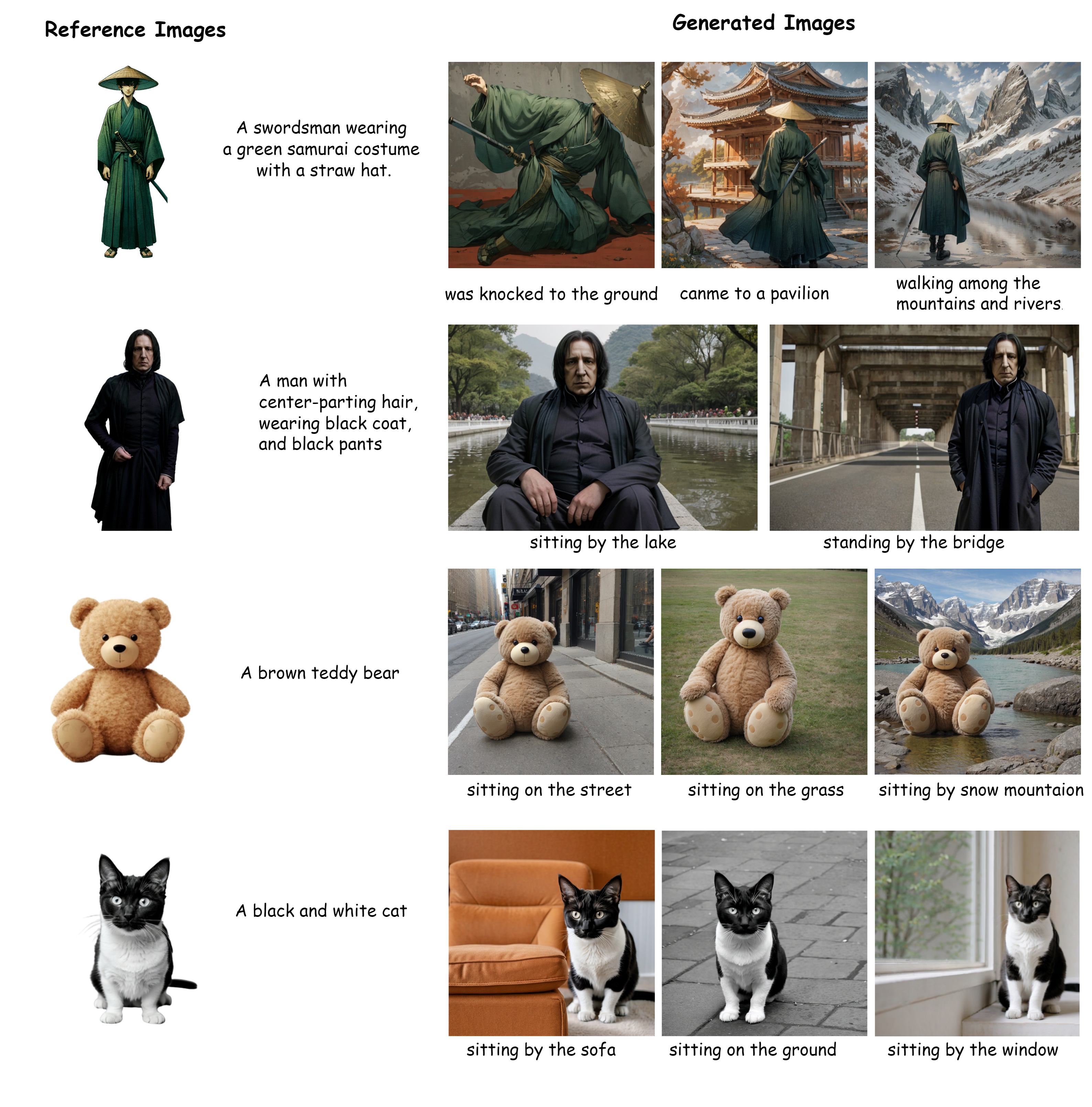} 
\caption{\label{fig:additional_result1} Additional qualitative results of single character. }
\end{figure*}

\begin{figure*} [thb]
\centering
\includegraphics[width=0.95\linewidth]{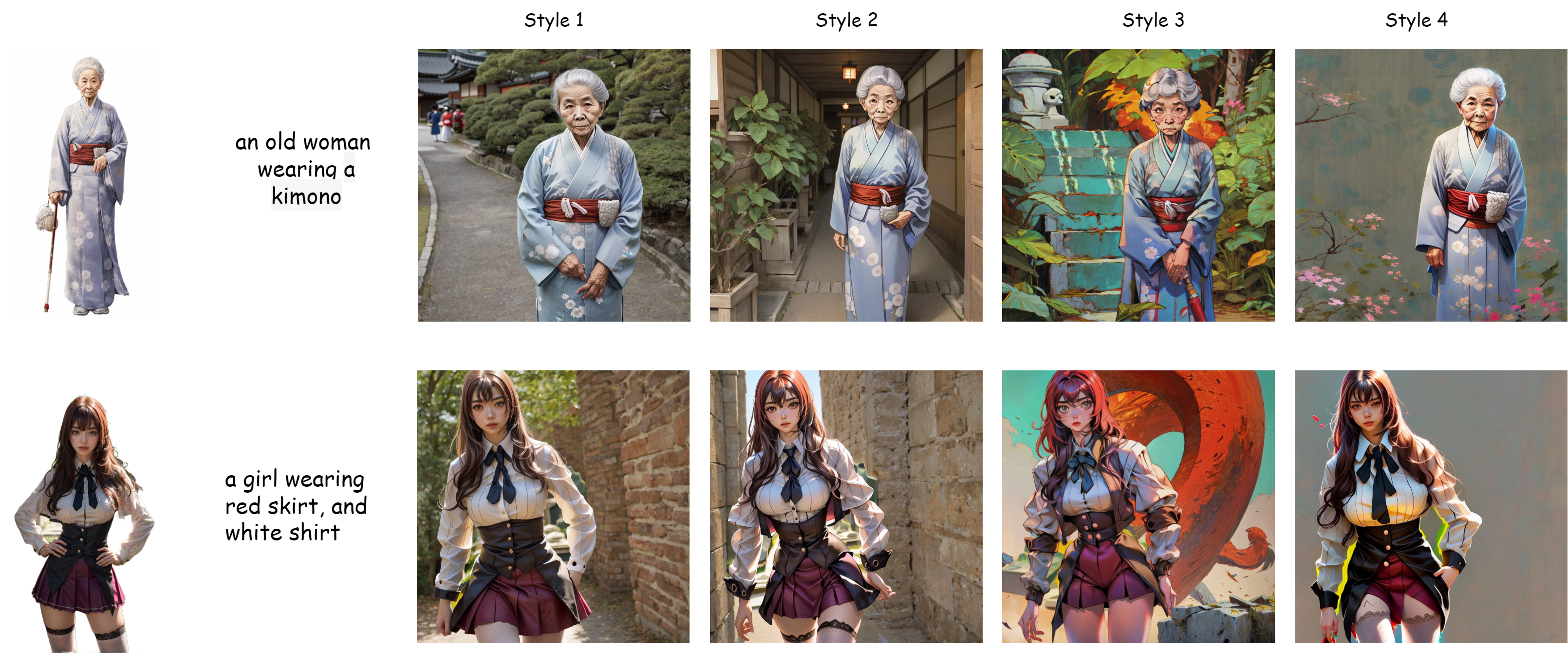} 
\caption{\label{fig:style_result} Additional qualitative results. }
\end{figure*}

In Fig. \ref{fig:additional_result2} and Fig. \ref{fig:additional_result1}, we present additional results showcasing the capabilities of Character-Adapter. The proposed framework demonstrates high-fidelity generation of consistent characters and objects across diverse scene contexts. Furthermore, we illustrate that Character-Adapter excels in scenarios involving multiple characters, maintaining high-fidelity of each character throughout the generated images. These results further validate the efficacy and versatility of our proposed approach. As shown in Fig. \ref{fig:style_result}, we also demonstrate the effectiveness of our approach across various stylized base models, thereby substantiating the generalizability of our proposed method.

\end{document}